\documentclass[letterpaper,10pt,conference]{ieeeconf}

\IEEEoverridecommandlockouts
\usepackage{cite}
\usepackage{graphicx}
\usepackage{amsmath,amssymb}
\usepackage{balance}
\usepackage{hyperref}
\graphicspath{{images/}}

\title{A Deployable Four-Finger Payload for Teleoperated Free-Flying
Manipulation with Astrobee}

\author{
William Su$^{1}$,
Jordan Kam$^{2}$,
Yunosuke Nakamura$^{3}$,
Yixiao Wang$^{3}$,
Jianshu Zhou$^{4}$,
and Masayoshi Tomizuka$^{3}$%
\thanks{$^{1}$Aerospace Engineering Program, University of California, Berkeley,
Berkeley, CA 94720, USA.}%
\thanks{$^{2}$Department of Aerospace Engineering, California Institute of Technology,
Pasadena, CA 91125, USA.}%
\thanks{$^{3}$Department of Mechanical Engineering, University of California, Berkeley,
Berkeley, CA 94720, USA.}%
\thanks{$^{4}$Department of Mechanical Engineering, National University of Singapore,
Singapore 117575.}%
}

\begin{document}
\maketitle

\begin{abstract}
This article presents a bimanual teleoperation pipeline and conceptual design of a deployable four-finger payload for intra-vehicular free-flyers. Future habitats in low-Earth orbit (LEO) will require systems to perform mundane tasks like cargo handling and maintenance during crewed and uncrewed periods. The gripper payload provides 17 manipulation degrees-of-freedom (DoF) through four independently actuated fingers on a linear rail system. To control it, a virtual reality (VR) device interface maps the human ground operator's hand motions to the finger pairs, their separation to the rail, and common wrist motion to Astrobee translation. We present the preliminary results of teleoperating Astrobee in a custom zero-gravity MuJoCo-based International Space Station (ISS) simulator through ten repeated trials of transporting a rigid ISS Cargo Transfer Bag (CTB). We measure task success, continuous contact retention, completion time, and cargo motion.

\end{abstract}

\section{Introduction}
\label{sec:introduction}

Near-term commercial low-Earth orbit destinations (CLDs) will require recurring habitat resupply, maintenance, and logistics management, including operations during periods with limited or no onboard crew \cite{graham2023cld, smith2021isaac}. Free-flying space robots could reduce the crew
time required for performing these routine tasks \cite{kam2025logistics, mckinley2021logistics}. Free-flyers such as the Japan Aerospace Exploration Agency's (JAXA) JEM Internal Ball Camera (Int-Ball) robot use this mobility for image capture and visual habitat inspection
\cite{mitani2019intball,hirano2025intball2}. While others like NASA's Astrobee free-flyer can also perform autonomous navigation through the ISS, docking, perching onto handrails, and modular payload integration increasing its capabilities~\cite{smith2016astrobee,barlow2018guestscience}. Astrobee's standard arm payload is a compliant perching arm was designed to securely grasp the handrails along the Kibo module of the ISS. However, this arm is limited in the range of dexterous tasks Astrobee can perform, even in free-flight
\cite{park2017developing}.

\begin{figure}[!t]
    \centering
    \IfFileExists{images/fig1.png}{%
        \includegraphics[width=\columnwidth]{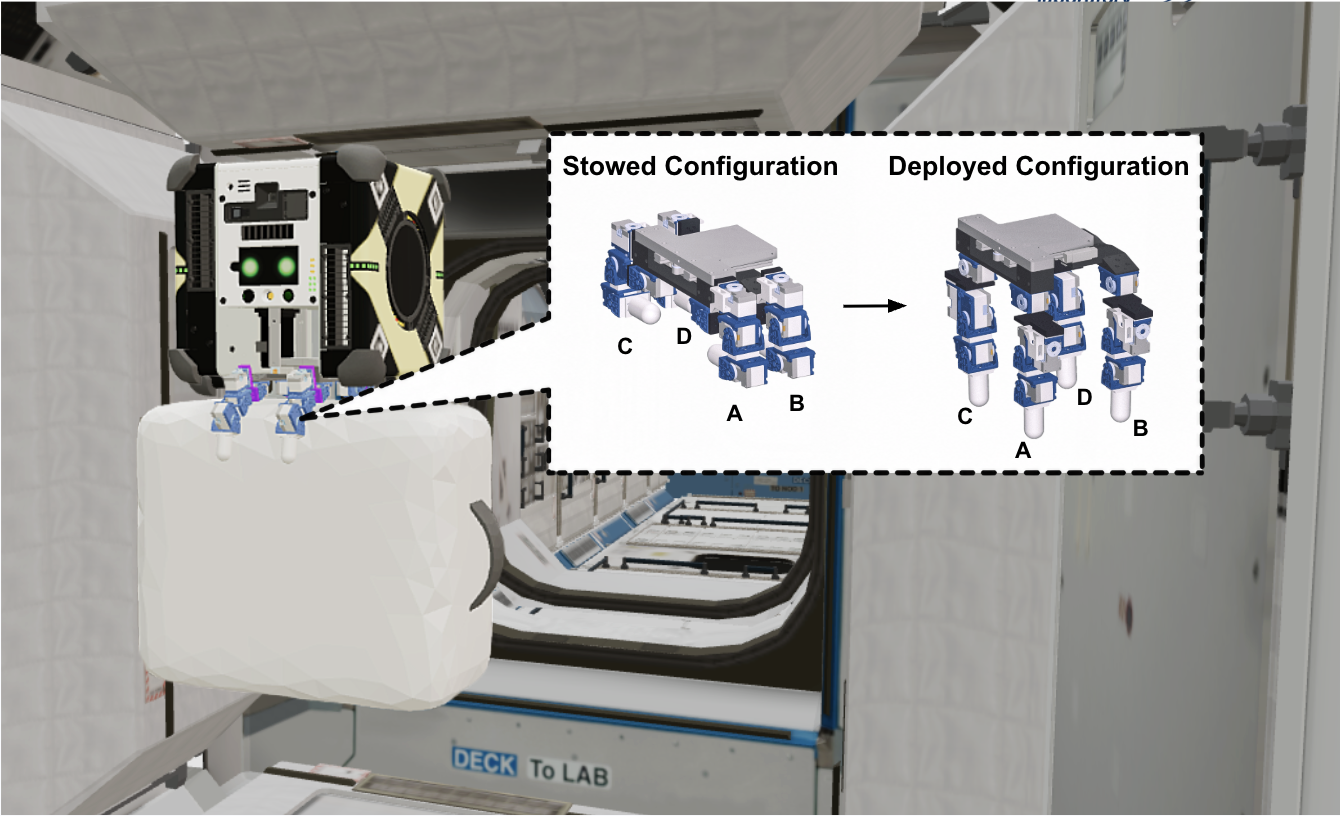}%
    }{%
        \fbox{\parbox[c][0.20\textheight][c]{0.92\columnwidth}{\centering
        Replace with the Astrobee payload and ISS cargo-scene figure.}}%
    }
    \caption{Astrobee equipped with the four-finger payload in (a) stowed and
(b) deployed configurations transporting a single ISS CTB in simulation.}
    \label{fig:astrobee_payload_iss}
\end{figure}

Several robotic gripper payloads have been developed in the past to extend Astrobee's manipulation capabilities. Gecko-inspired adhesives have been used to support attachment to smooth surfaces inside the ISS \cite{chen2022testing}. Hybrid rigid-soft pneumatic grippers have been conceptually explored to grasp onto complex geometries using Astrobee
\cite{kam2025underactuated}. Other concepts have explored how pneumatic actuation can enable dexterous two-finger, 6-DOF grippers
for both perching and continuous in-hand
manipulation \cite{su2026dexcoastrobee, zhou2025dexco}. These grippers have extended Astrobee from solely attachment grasping towards fine object manipulation.  Dual-arm systems add reach and independent contact placement, but increase stowed volume and require coordination of two arm chains with the free-flying base \cite{su2026dualarm}. For objects that Astrobee can approach directly, vehicle motion provides gross positioning without the need for extended arm reach. To address this, we present a deployable four-finger dexterous payload and teleoperation pipeline for controlling this gripper. Our contributions include:

\begin{enumerate}
    \item Conceptual payload and gripper design of a deployable four-finger rail actuation system.
    \item  A Quest-based virtual reality (VR) interface to control the fingers, pair-spacing rail, and robot motion concurrently.
    \item An evaluation of high dexterity teleoperation with Astrobee for rigid-bodied ISS Cargo Transfer Bags (CTBs). 
\end{enumerate}

\begin{figure*}[!t]
    \centering
    \includegraphics[width=\textwidth]  {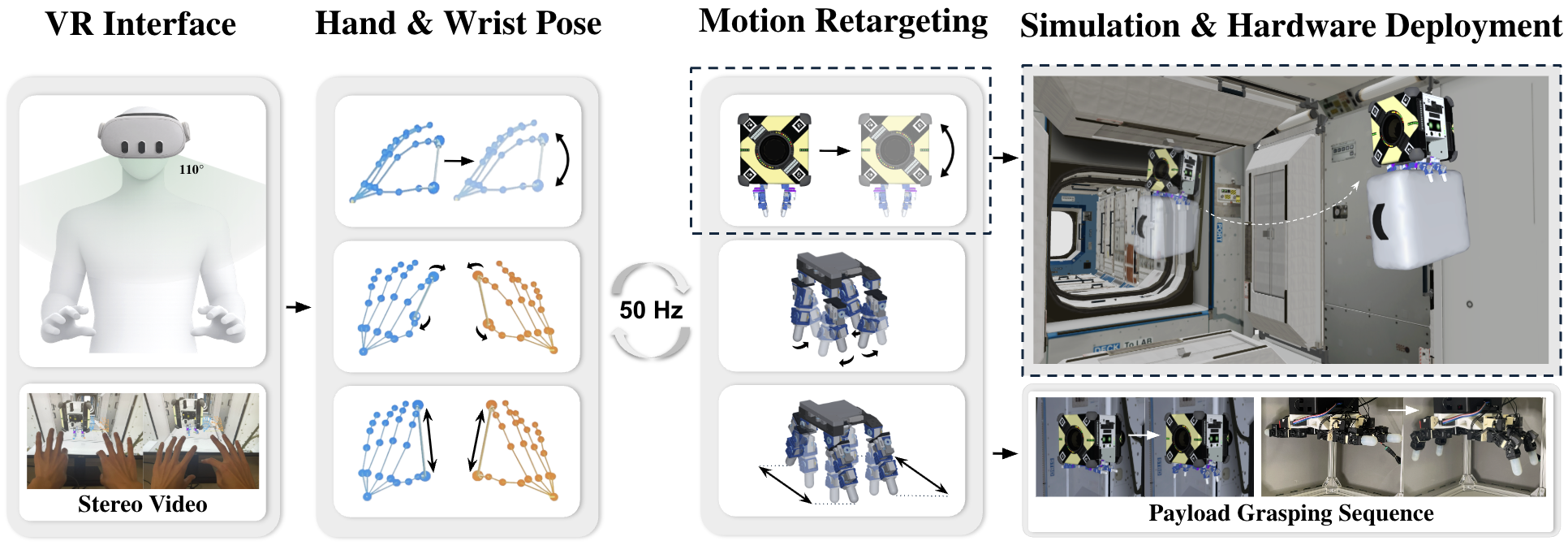}%
    
    \caption{Bimanual teleoperation pipeline overview. Calibrated hand features command the two finger pairs and spacing rail, while common wrist translation commands Astrobee motion. All three commands are generated concurrently.}
    \label{fig:teleop_pipeline}
\end{figure*}

\section{Methodology}

The payload uses four multi-DOF fingers inspired by the Low-cost, Efficient, and Anthropomorphic (LEAP) Hand \cite{shaw2023leap}. Fig.~\ref{fig:astrobee_payload_iss} outlines the four-finger payload attached to Astrobee grasping an ISS CTB. Fingers A and C are fixed to the structure, while fingers B and D share a carriage driven along a 50-mm linear rail, providing 16 finger DoF and the rail adds one pair-spacing DoF\cite{su2026dualopposition}. A vertical stage stows the mechanism
within the combined envelope of Astrobee's two lower 1-U payload bays and deploys the payload below the vehicle for manipulation. The present study evaluates the integrated system in simulation; hardware validation is in preparation and not reported here. Its commanded manipulation configuration follows:

\begin{equation}
\mathbf{q}=
\begin{bmatrix}
\mathbf{q}_{A}^{\mathsf T} &
\mathbf{q}_{B}^{\mathsf T} &
\mathbf{q}_{C}^{\mathsf T} &
\mathbf{q}_{D}^{\mathsf T} & q_r
\end{bmatrix}^{\mathsf T}\in\mathbb{R}^{17},
\label{eq:platform_configuration}
\end{equation}

where $\mathbf{q}_{i}\in\mathbb{R}^{4}$ contains the
metacarpophalangeal (MCP) flexion and abduction/adduction, proximal
interphalangeal (PIP), and distal interphalangeal (DIP) angles of finger $i$.
The scalar $q_r$ is the pair-spacing displacement. The deployment coordinate
is separate from these 17 manipulation coordinates.

\subsection{Teleoperation Pipeline}

Fig.~\ref{fig:teleop_pipeline} summarizes the teleoperation pipeline for the
dexterous payload. The VR interface tracks the wrist and the distal-joint and fingertip positions of the thumb and index finger on each hand. The relative thumb--index geometry specifies the desired contact configuration for the corresponding eight-DoF finger pair. Inter-wrist distance controls the pair-spacing rail, while common wrist translation controls Astrobee motion. An optimization-based retargeting step maps these task-space targets to commands for the 16 finger joints.

\begin{figure*}[!t]
    \centering
    \includegraphics[width=\textwidth]  {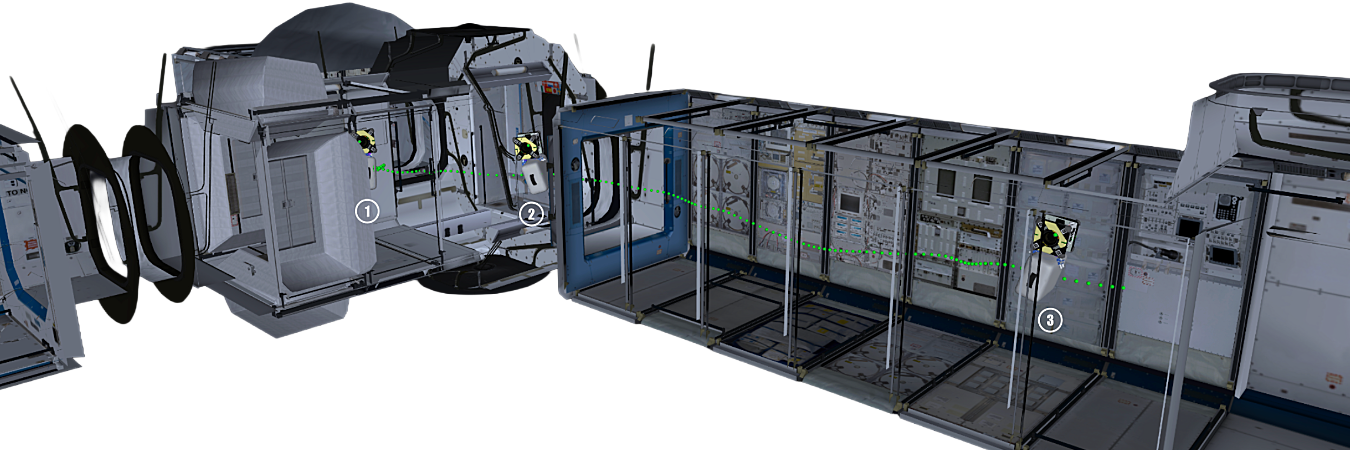}%
    
    \caption{Recorded cargo-transport trial. The sequence contains (1) cargo
    acquisition and translation, (2) a commanded turn, and (3) arrival. The trajectory
    is the cargo center of mass expressed in the Astrobee body frame.}
    \label{fig:cargo_transport}
\end{figure*}

For hand $s\in\{L,R\}$, let
$\mathbf{q}_s\in\mathbb{R}^{8}$ contain the joints of its assigned robot
pair. Let $\mathbf{p}_{s,i}(\mathbf{q}_s)$ and
$\mathbf{a}_{s,i}(\mathbf{q}_s)$ denote the fingertip and distal-joint
positions of finger $i\in\{1,2\}$ in the payload frame. The pair geometry is

\begin{equation}
\begin{aligned}
\mathbf{d}_s
&=\mathbf{p}_{s,2}-\mathbf{p}_{s,1},&
\ell_s
&=\|\mathbf{d}_s\|,\\
\hat{\mathbf{d}}_s
&=\mathbf{d}_s/\ell_s,&
\mathbf{c}_s
&=\tfrac{1}{2}(\mathbf{p}_{s,1}+\mathbf{p}_{s,2}),\\
\mathbf{u}_{s,i}
&=\frac{\mathbf{p}_{s,i}-\mathbf{a}_{s,i}}
{\|\mathbf{p}_{s,i}-\mathbf{a}_{s,i}\|},&
i&\in\{1,2\}.
\end{aligned}
\label{eq:pair_features}
\end{equation}

Here, $\ell_s$ is the fingertip separation,
$\hat{\mathbf{d}}_s$ is the direction from the first fingertip to the second,
$\mathbf{c}_s$ locates the pair, and $\mathbf{u}_{s,i}$ gives the orientation
of each distal link. The vector
$\mathbf{u}_s=[\mathbf{u}_{s,1}^{\mathsf T},
\mathbf{u}_{s,2}^{\mathsf T}]^{\mathsf T}\in\mathbb{R}^{6}$ stacks these two
directions. The same features are extracted from the tracked thumb and index finger in
the calibrated wrist frame. The pinch-to-open calibration maps them to the
robot-space targets
$(\bar{\ell}_s,\bar{\mathbf{d}}_s,\bar{\mathbf{c}}_s,
\bar{\mathbf{u}}_s)$. The overbars denote mapped robot targets. Calibration also supplies a posture reference consistent
with the commanded opening.

Let $\widetilde{\mathbf{q}}_s$ denote joint coordinates normalized by their
respective ranges. The posture reference and preceding command use the same
normalization. Let
$\|\mathbf{x}\|_{\rho}^{2}=\sum_k\rho(x_k^2)$ denote the componentwise
soft-$L_1$ cost. The command at control step $t$ is

\begin{equation}
\begin{aligned}
\mathbf{q}_{s,t}^{*}
={}&\operatorname*{argmin}_{
\mathbf{q}_s^{\min}\leq\mathbf{q}_s\leq\mathbf{q}_s^{\max}}
\frac{1}{2}\Bigg[
\left\|\frac{\ell_s-\bar{\ell}_s}{\sigma_{\ell}}\right\|_{\rho}^{2}
+\left\|\frac{w_d(\hat{\mathbf{d}}_s-\bar{\mathbf{d}}_s)}
{\sigma_d}\right\|_{\rho}^{2}\\
&+\left\|\frac{\mathbf{c}_s-\bar{\mathbf{c}}_s}
{\sigma_c}\right\|_{\rho}^{2}
+\left\|\frac{w_f(\mathbf{u}_s-\bar{\mathbf{u}}_s)}
{\sigma_f}\right\|_{\rho}^{2}\\
&+\left\|w_p(\widetilde{\mathbf{q}}_s-
\widetilde{\mathbf{q}}_s^{\mathrm{ref}})\right\|_{\rho}^{2}
+\left\|w_t(\widetilde{\mathbf{q}}_s-
\widetilde{\mathbf{q}}_{s,t-1})\right\|_{\rho}^{2}
\Bigg].
\end{aligned}
\label{eq:retargeting}
\end{equation}

The first four terms track the commanded span, pair direction, pair center,
and distal-link directions. The posture term resolves kinematic redundancy
toward the calibrated reference, while the temporal term suppresses abrupt
changes in the range-normalized joint coordinates.

The geometric scales are $\sigma_{\ell}=1.5$~mm,
$\sigma_c=25$~mm, and $\sigma_d=\sigma_f=0.15$, with
$w_d=0.35$, $w_f=0.10$, $w_p=3.0$, and $w_t=0.03$. The length scales express
span and center errors in comparable dimensionless units. The direction terms
subtract unit vectors, so $\sigma_d$ and $\sigma_f$ are dimensionless chordal
scales; for small errors, 0.15 corresponds approximately to 0.15~rad. These
are fixed controller settings for all evaluation trials. This objective follows geometry-based retargeting
\cite{handa2020dexpilot,qin2023anyteleop}. At each control step, forward
kinematics evaluates the current robot geometry and analytic geometric
Jacobians describe how it changes with the eight joint angles. Each pairwise solve uses at most 20 least-squares function
evaluations per control step.

The other two channels bypass the finger optimizer. Calibrated inter-wrist
distance maps directly to the pair-spacing rail. Coordinated wrist translation
commands Astrobee through a virtual sphere centered at the neutral left-wrist
position. The left wrist sets the travel direction after leaving the neutral region, and its displacement beyond the boundary scales
the requested speed up to 0.20~m/s. Astrobee turns toward the
requested direction while translating, subject to bounded acceleration and
jerk. Returning both wrists to the neutral region commands smooth braking.

\subsection{ISS Simulation}

We evaluate the gripper and teleoperation pipeline in a custom zero-gravity
ISS simulation implemented in MuJoCo
\cite{todorov2012mujoco,kam2025towards}. Physics is integrated at 1~kHz
beneath the 50-Hz teleoperation loop. Astrobee and the CTB
\cite{morton2025deformable} are modeled as free rigid bodies; a bounded wrench tracks the commanded vehicle motion, while the CTB moves only through simulated contact as seen in Fig.~\ref{fig:cargo_transport}. The CTB transport task contains four phases. The user first flexes both finger pairs to acquire the CTB. Common wrist motion then translates Astrobee forward along the first route segment, changes its heading through a turn, then moves the CTB from the Kibo Module inside the ISS to a connected destination module. The fingers remain under retargeting control throughout the task.

The teleoperation pipeline records the hand command, robot joint state,
Astrobee pose, CTB pose, and an opposing-contact indicator at every
50-Hz control step. A contact is confirmed when the CTB contacts
fingertips from both opposing sides of the finger array. CTB position in
the instantaneous Astrobee body frame is

\begin{equation}
{}^{b}\mathbf{p}_{c}(t)=
\left(\mathbf{R}_{b}^{W}(t)\right)^{\mathsf{T}}
\left(\mathbf{p}_{c}^{W}(t)-\mathbf{p}_{b}^{W}(t)\right),
\label{eq:relative_cargo}
\end{equation}

where $\mathbf{p}_{b}^{W}$ and $\mathbf{p}_{c}^{W}$ are the world-frame
Astrobee and CTB positions, respectively, and
$\mathbf{R}_{b}^{W}\in\mathrm{SO}(3)$ maps body-frame vectors into the
world frame. For each trial, let $t_0$ denote the first sample in its
first continuous confirmed-contact interval, and let
$\{t_k\}_{k=1}^{N}$ denote all samples in that interval. We quantify
CTB translational deviation relative to the vehicle as

\begin{equation}
e_{p,\mathrm{RMS}}=
\sqrt{\frac{1}{N}\sum_{k=1}^{N}
\left\|{}^{b}\mathbf{p}_{c}(t_k)
-{}^{b}\mathbf{p}_{c}(t_0)\right\|_2^2}.
\label{eq:cargo_rms}
\end{equation}

This metric removes commanded Astrobee translation and rotation, and
measures CTB translational motion relative to the vehicle during the
confirmed-contact interval.

\section{Results}

All ten live teleoperation trials established a first continuous interval of confirmed opposing CTB contact and executed the intended acquisition, translation, and right-turn maneuver. The median recorded teleoperation duration, measured after calibration, was 77.5~s (range: 67.3--97.5~s). The CTB center of mass underwent a median net world-frame displacement of 9.30~m (range: 7.87--9.46~m). To quantify cargo motion relative to the free flyer, RMS translational displacement from the CTB position in the Astrobee body frame at first confirmed contact was computed for each trial.  The median RMS deviation across the ten trials was 35.6~mm (range: 16.4--79.0~mm), and the median peak deviation was 67.7~mm (range: 26.8--176.6~mm). The corresponding confirmed-contact interval lasted 75.1~s (range: 65.7--94.5~s). The result indicates that the
integrated hand and base controller can retain the simulated cargo while the operator changes the free-flyer's trajectory.

\section{Conclusion \& Future Work}

This article presented the conceptual design and evaluation of a deployable four-finger payload and bimanual teleoperation pipeline using Astrobee. The VR interface concurrently maps hand articulation to two finger pairs, wrist separation to the payload rail, and common wrist motion to the free-flying base. An ISS-like simulation showed how this pipeline can be used to teleoperate Astrobee in transporting cargo. Future work will characterize the completed payload hardware,
evaluate repeated operators and tasks, incorporate collision-aware
Astrobee control, and transfer the teleoperation pipeline to hardware.
Recorded teleoperated trajectories will also provide demonstrations
for subsequent imitation-learning studies using the same embodiment.

\bibliographystyle{IEEEtran}
\bibliography{references}

\end{document}